\documentclass[11pt]{article}

\usepackage[preprint]{acl}

\usepackage{times}
\usepackage{latexsym}
\usepackage{amsmath,amssymb,booktabs}
\usepackage{makecell}
\usepackage[T1]{fontenc}

\usepackage[utf8]{inputenc}

\usepackage{microtype}
\usepackage{graphicx}
\usepackage{subcaption}
\usepackage{tabularx}

\usepackage{inconsolata}
\usepackage{booktabs,multirow}

\title{What LLMs Do in the Shadows: Reading Context Reliance from the Unembedding Dark Space}

\author{Alexandra Kuleshova, Andrey Volodichev, Daria Kotova, Alexey Zaytsev\\
  Applied AI Institute \\
  \texttt{bazarovaai.239@gmail.com} \\}

\begin{document}
\maketitle
\begin{abstract}
Detecting faithfulness hallucinations in large language models (LLMs) is crucial for their safe deployment in high-stakes domains. Recent approaches address this by contrasting contextual against parametric knowledge utilization, yet measuring context reliance typically demands noticeable overhead, such as materializing attention maps, running additional forward passes, or training auxiliary probes. In this work, we show that context reliance can be estimated by projecting hidden states onto the \emph{dark space}~---~directions that compose weakly with the unembedding matrix. Beyond its recognized function as an initial-token attention sink, we show that this subspace actively mediates context processing, and that its share of hidden-state norm~---~termed the \emph{Dark Matter Ratio} ($\operatorname{DMR}$)~---~provides a direct score of context utilization. Pairing $\operatorname{DMR}$ with a parametric knowledge score yields \textsc{DarkLens}, a lightweight faithfulness detector operating strictly within a single forward pass. Evaluated across three QA and summarization benchmarks on six open-weight LLMs (4B--13B parameters), \textsc{DarkLens} matches or outperforms a wide range of baselines without requiring extra forward passes or probe training.\footnote{Our code is available at the following \href{https://anonymous.4open.science/r/dark_lens-A728/}{link}.}
\end{abstract}

\section{Introduction}
\label{sec:intro}

In context-grounded tasks such as retrieval-augmented generation and summarization, large language models frequently generate \emph{faithfulness hallucinations}~---~claims that contradict or cannot be verified against the source text~\citep{niu2024ragtruth,chuang2024lookback}.
Detecting such errors at inference time is crucial for reliable deployment, particularly in high-stakes domains such as medicine, law, or finance where adherence to source documents is non-negotiable.

Prior work indicates that faithfulness hallucinations frequently stem from conflicts between a model's internal knowledge and the provided context~\citep{xu2024knowledge}.
In particular, \citet{sun2025redeep} demonstrate that such failures occur when models prioritize parametric memory over contextual evidence, showing that hallucinations can be detected by contrasting a context utilization score against a parametric reliance score.
However, current approaches introduce substantial overhead, relying on attention-map materialization~\citep{sun2025redeep}, counterfactual forward passes~\citep{yeh2026lumina}, or auxiliary probe training~\citep{wang2025seredeep}. 

In this paper, we show that context reliance can be estimated directly from residual hidden states by projecting them onto the \emph{dark space} of the unembedding matrix $W_U$~\citep{cancedda2024spectral}~---~directions that compose weakly with $W_U$ and minimally affect vocabulary logits. Contrary to the hypothesis that this subspace primarily serves to form initial-token attention sinks, we find that dark components at non-BOS positions actively mediate context processing and retrieval. Grounded in this finding, we introduce the \emph{Dark Matter Ratio} ($\operatorname{DMR}$), which measures the fraction of hidden-state norm residing in the dark space to serve as an attention-free context reliance score.

Building on $\operatorname{DMR}$, we introduce \textsc{DarkLens}, a lightweight faithfulness detector that pairs this context utilization signal with the parametric knowledge score of \citet{sun2025redeep}.
Because both quantities are computed solely from residual hidden states and static projections, \textsc{DarkLens} operates in a single forward pass without requiring attention-map materialization or probe training.

Our main contributions are:
\begin{itemize}
    \item We examine the dark components of hidden states~---~directions that compose weakly with the unembedding matrix $W_U$~---~and show that, contrary to prior hypotheses that this subspace primarily serves to form initial-token attention sinks, they actively mediate context processing across non-BOS positions;
    \item Based on these findings, we propose the Dark Matter Ratio ($\operatorname{DMR}$), an attention-free context utilization score defined as the fraction of hidden-state norm residing in the dark subspace, which avoids attention map materialization and extra forward passes;
    \item We introduce \textsc{DarkLens}, a lightweight detector combining $\operatorname{DMR}$ with a parametric knowledge signal in a single forward pass. Evaluated across three benchmarks on six open-weight LLMs (4B--13B parameters), \textsc{DarkLens} achieves performance on par with or superior to a wide range of baselines.
\end{itemize}

\section{Background}
\label{sec:background}

In autoregressive transformers, hidden representations in the residual stream are mapped to vocabulary logits via the unembedding matrix $W_U \in \mathbb{R}^{d_{\mathrm{vocab}} \times d_{\mathrm{model}}}$:
\begin{equation}
z^{(\ell)}_t = W_U h^{(\ell)}_t , \qquad p^{(\ell)}_t = \mathrm{softmax}\!\left(z^{(\ell)}_t\right).
\label{eq:unembedding}
\end{equation}
While popular interpretability techniques such as the logit lens~\citep{logitlens,belrose2023eliciting} use this mapping to inspect representations across layers, they inherently overlook components that have low composition with the unembedding matrix and thus have little effect on output logits.

\citet{cancedda2024spectral} formalized these weakly composing directions via the singular value decomposition $W_U = U \Sigma V^{\top}$, defining the \textit{dark space} as the subspace spanned by the right singular vectors $V_{\mathrm{dark}} \in \mathbb{R}^{d_{\mathrm{model}} \times d_{\mathrm{dark}}}$ corresponding to the $5\%$ of the smallest singular values.
Projections onto this subspace induce negligible logit variation, rendering them largely invisible to vocabulary-level readouts.
Using the orthogonal projection matrix $P_{\mathrm{dark}} = V_{\mathrm{dark}} V_{\mathrm{dark}}^{\top}$, any hidden state $h$ decomposes into:
\begin{equation}
h = h_{\perp} + h_{\mathrm{d}},
\label{eq:dark-decomposition}
\end{equation}
where $h_{\mathrm{d}} = P_{\mathrm{dark}} h$ is the dark component and $h_{\perp} = (I - P_{\mathrm{dark}}) h$ is the vocabulary-visible component.
\citet{cancedda2024spectral} showed that the initial (BOS) token resides almost entirely in this dark space ($h \approx h_{\mathrm{d}}$), allowing it to act as an attention sink~\citep{gu2025attention} without distorting vocabulary logits.
They hypothesized that the dark space primarily enables such BOS sinks, leaving its role at subsequent tokens unestablished.
In this work, we examine dark representations at non-BOS positions across intermediate layers and explore their connection to context utilization.

\section{Empirical Study}\label{sec:empirical-study}
In every experiment below we leave the residual stream at the BOS token unchanged, so the effects we measure are separate from the attention-sink role that \citet{cancedda2024spectral} attributes to the dark subspace at that position. 

\subsection{Synthetic Induction}\label{sec:synthetic-induction}

\begin{figure*}[tbp]
  \centering
  \includegraphics[width=\linewidth]{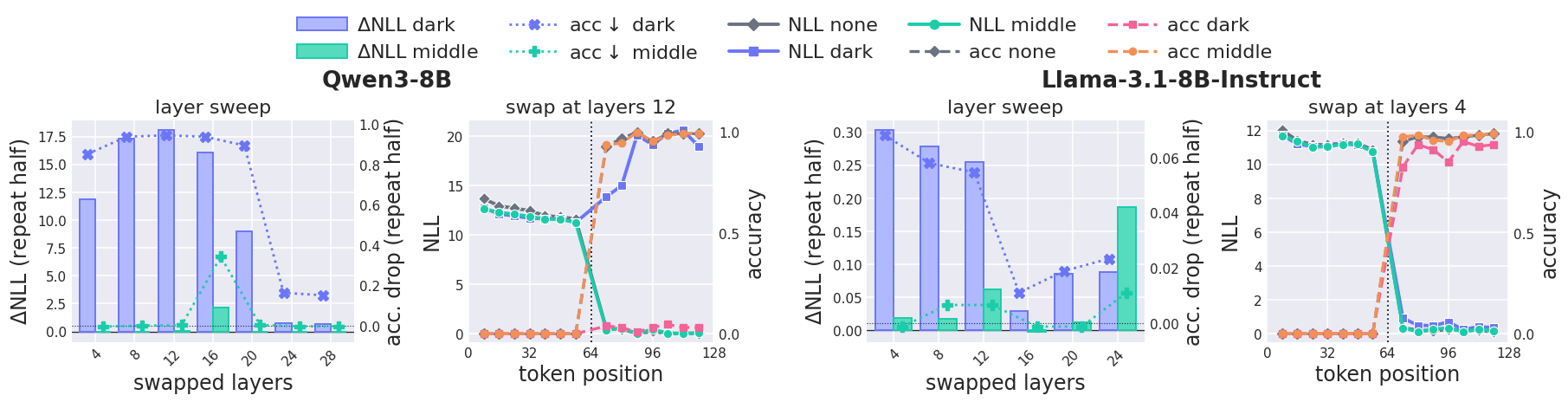}
  \caption{Synthetic induction residual swaps on $[x_1{\ldots}x_T,\,x_1{\ldots}x_T]$.
  Each intervention covers a window of four consecutive layers.
  The first and third panels plot $\Delta$NLL on the repeated half against the start layer of that window (\texttt{dark} vs.\ mid-spectrum control); the dotted line is the average accuracy drop on the second half of the sequence.
  The second and fourth plot NLL (solid) and accuracy (dashed) against token position at the peak window; the vertical line marks the start of the repeat.
  Dark swaps impair copying far more than matched mid-spectrum swaps.}
  \label{fig:synthetic_dm_swap}
\end{figure*}

\paragraph{Task setup.} We first analyze the functions of the dark components in a synthetic copying scenario. We generate $N = 128$ sequences of random tokens, where each synthetic sequence has the form $[x_1,\ldots,x_T,\,x_1,\ldots,x_T]$, $T = 64$: the first half is a string of randomly sampled tokens, and the second half repeats it.
Models normally show a sharp NLL drop (and near-perfect accuracy) once the repeated half begins, as they begin to copy from the matched earlier positions~\cite{stolfo2024confidence}.  

\paragraph{Intervention.} We intervene in the manner of \citet{cancedda2024spectral}.
Within each window of four consecutive layers\footnote{We omit the first and last layer windows from the visualization, as interventions in these boundary layers cause substantially more severe model degradation (see Appendix\ref{sec:app-synth-induction}).}, we exchange the dark components across synthetic samples at the same token position.
Specifically, for a target state $h$ and a donor state $h'$ at the same position, we replace the target's dark component with that of the donor: $\hat h = h_{\perp} + h'_{\mathrm{d}},$
and rescale $\hat h$ to the original norm $\|h\|_2$ to keep activation scale unchanged.

\begin{figure}[tbp]
  \centering
  \includegraphics[width=\linewidth]{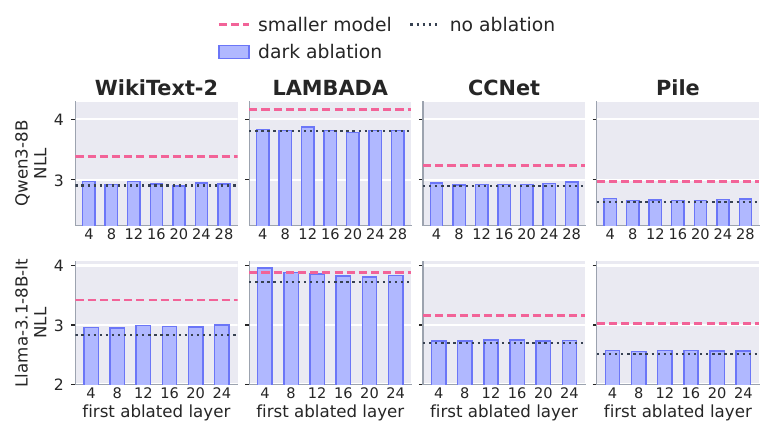}
  \caption{Language-modeling NLL under dark vs.\ mid-spectrum residual ablation.
  Each intervention covers a window of four consecutive layers; bars show NLL against the first layer of that window.
  Solid grey: unablated large model; dashed pink: smaller reference model (Qwen3-1.7B and Llama-3.2-1B-Instruct).}
  \label{fig:nll_general}
\end{figure}

\paragraph{Controls.} To verify that any drop in copying does not merely reflect general language-modeling degradation, we measure NLL on four standard corpora under dark interventions (Figure~\ref{fig:nll_general}).
Across most layer windows, dark interventions maintain NLL close to the unablated model and below that of a smaller reference model (Qwen3-1.7B, Llama-3.2-1B-Instruct), indicating that general language-modeling capabilities are largely preserved.

To further ensure that the copying impairment cannot be explained by general damage, we compare against an NLL-matched control: for each layer window, we measure WikiText-2 NLL under the dark swap, then find a mid-spectrum subspace of $W_U$ (with rank $\ge d_{\text{dark}}$) that produces the same NLL increase (Table~\ref{tab:wikitext-middle-ranks}).
As elsewhere, we never modify the representation of the first token.

\paragraph{Results.} Figure~\ref{fig:synthetic_dm_swap} presents the results for Qwen3-8B and Llama-3.1-8B-Instruct.
While the magnitude of the effect is model-dependent, dark-matter swaps consistently impair copying across all evaluated windows.
On Qwen3-8B, dark interventions in windows starting at layers 4--20 reduce copying accuracy on the repeated half nearly to zero (dropping by up to 95\%), while the matched mid-spectrum control leaves it largely intact.
On Llama-3.1-8B-Instruct, the accuracy drop is more modest but still exceeds that of the control; only in layers 24--27 does the mid-spectrum swap cause a larger $\Delta$NLL (likely due to the high rank needed to match general NLL; Table~\ref{tab:wikitext-middle-ranks}), yet copying accuracy still degrades more under the dark intervention.
Overall, dark swaps consistently disrupt copying, especially in early-to-mid layers, suggesting that dark directions participate in context retrieval.

\subsection{Dark Intervention: Contextual vs Parametric}\label{sec:ctx-vs-param}

\paragraph{Task setup.} To test whether dark components specifically facilitate context utilization in natural language tasks, we compare model behavior across contextual and parametric benchmarks.
For contextual reasoning, we use NQ-Swap~\cite{longpre-etal-2021-entity} and ConflictQA~\cite{xie2024adaptive}, where the provided passage contains an injected fact that directly conflicts with real-world knowledge; answering faithfully thus forces the model to rely on the external context rather than its stored memory (Table~\ref{tab:nqswap-conflictqa-examples}).
In contrast, the parametric datasets (Table~\ref{tab:factual-qa-examples})~---~Cities, Sports, and Celebrity~\cite{pmlr-v267-jin25f}~---~provide no context at all, consisting of isolated binary yes/no questions that probe factual recall directly from the model's weights.

\paragraph{Intervention.} As in Section~\ref{sec:synthetic-induction}, we apply interventions across windows of four consecutive layers.
Because natural sequences vary in length across examples, we permute the dark components across positions within each individual sequence rather than across sequences.
Formally, for a sequence of hidden states $(h_1, \ldots, h_T)$ at each intervened layer within the window, let $\pi$ be a random permutation of the non-initial token indices $\{2, \ldots, T\}$.
For each $t \in \{2, \ldots, T\}$, we construct
\[
\tilde h_t = h_{t,\perp} + h_{\pi(t),\mathrm{d}}, \qquad \hat h_t = \|h_t\|_2 \, \frac{\tilde h_t}{\|\tilde h_t\|_2},
\]
while leaving the initial token unmodified ($\hat h_1 = h_1$).

\paragraph{Controls.} As a control, we apply the same permutation within a subspace spanned by mid-spectrum right singular vectors, choosing its dimension such that the intervention matches the NLL degradation induced by the dark permutation on WikiText-2 (Table~\ref{tab:wikitext-middle-perm-nll-match}).

\paragraph{Results.} We evaluate the target preference rate: the fraction of examples where the target continuation (context-aligned for contextual tasks, factually correct for parametric ones) achieves lower NLL than the alternative.
Table~\ref{tab:subspace-hit-rates} presents the results for the intervention on layers 20--23 (see Appendix~\ref{sec:app-results} for other layer windows).
For Qwen3-8B, the dark intervention sharply impairs contextual preference (dropping by up to 75\% relative on NQ-Swap) while leaving parametric accuracy almost intact.
By contrast, the mid-spectrum control only slightly degrades contextual preference, but noticeably damages parametric performance (by up to 17\%).
For Llama-3.1-8B-Instruct the effect is milder but still present: the dark intervention consistently reduces contextual preference while leaving parametric performance essentially unchanged.
These patterns suggest that dark subspace disruption primarily weakens context reliance, with comparatively little impact on parametric recall.

\begin{table}[tbp]
  \centering
  \small
  \setlength{\tabcolsep}{2pt}
  \begin{tabular}{llccc}
  \toprule
  & & & \multicolumn{2}{c}{L20--23} \\
  \cmidrule(lr){4-5}
  Model & Dataset & No ablation & Middle & Dark \\
  \midrule
  \multirow{5}{*}{Qwen3-8B}
    & NQ-Swap    & 0.88 & 0.81 & 0.22 \\
    & ConflictQA & 0.98 & 0.99 & 0.54 \\
    \cmidrule(lr){2-5}
    & Cities     & 1.0  & 0.83 & 0.99 \\
    & Sports     & 0.95 & 0.81 & 0.93 \\
    & Celebrity  & 0.96 & 0.80 & 0.95 \\
  \addlinespace
  \midrule
  \multirow{5}{*}{Llama-3.1-8B-IT}
    & NQ-Swap    & 0.83 & 0.87 & 0.65 \\
    & ConflictQA & 0.96 & 0.96 & 0.92 \\
    \cmidrule(lr){2-5}
    & Cities     & 1.0  & 1.0  & 1.0  \\
    & Sports     & 0.95 & 0.95 & 0.94 \\
    & Celebrity  & 0.96 & 0.96 & 0.95 \\
  \bottomrule
  \end{tabular}
  \caption{Share of examples on which the target continuation has strictly lower mean NLL than the alternative. Datasets above the rule are contextual (NQ-Swap, ConflictQA); those below are parametric (Cities, Sports, Celebrity). Middle and dark permute layers 20--23. Dark subspace dimension 205 (last 5\% of right singular vectors of $W_U$).}
  \label{tab:subspace-hit-rates}
  \end{table}

\begin{figure}[tbp]
    \centering
    \includegraphics[width=\linewidth]{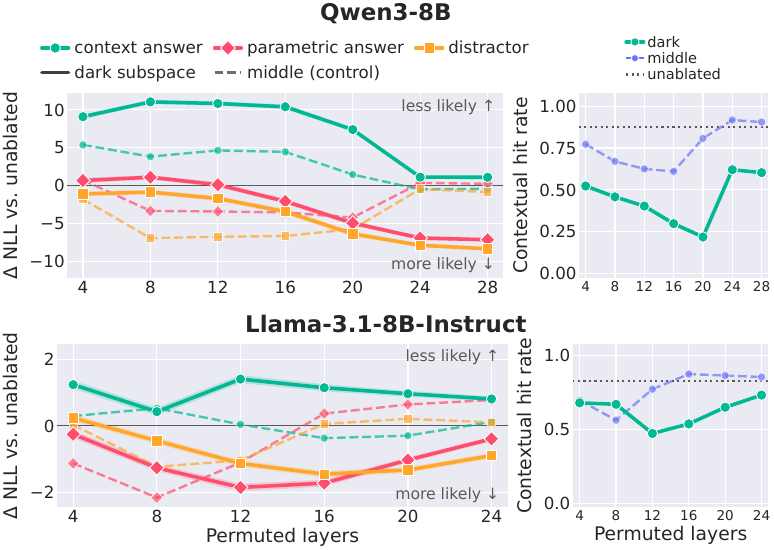}
    \caption{NQ-Swap answer NLL under dark vs.\ mid-spectrum residual permutation.
    $\Delta$NLL is relative to the unablated model (positive: less likely).
    Solid: dark subspace; dashed: matched mid-spectrum control.
    Context, parametric, and distractor answers are scored separately.}
    \label{fig:nq-swap-perm-nll}
\end{figure}

  Figure~\ref{fig:nq-swap-perm-nll} provides a layer-wise breakdown on NQ-Swap, also tracking a plausible \textit{distractor} (an entity appearing neither in the passage nor in parametric memory).
  Because state permutations disrupt positional cues, the mid-spectrum control can also reduce context hit rates in early layers for Llama (even exceeding the dark intervention) and across early-to-mid layers for Qwen.
  Across most windows, however, the dark intervention tends to impair context reliance more substantially, leading to larger NLL penalties on context-aligned answers and deeper drops in hit rate.
  Notably, under dark interventions, parametric and distractor NLLs decrease together almost identically.
  This suggests that the dark subspace does not specifically promote parametric facts, but rather facilitates context utilization: disrupting it appears to cause the model to fall back on general token plausibility without selectively favoring the parametric answer over an arbitrary distractor.

\subsection{Dark Intervention: Modulating Context Use}

\paragraph{Task setup.} If dark components mediate context use, modulating their magnitude may steer context reliance.
To test this hypothesis, we evaluate models on the contextual conflict benchmarks NQ-Swap and ConflictQA, scoring answers by NLL to measure the contextual hit rate as a function of intervention strength $\alpha$.

\paragraph{Intervention.} We intervene on layers 20--23 using the rank-$205$ dark subspace ($d_{\text{dark}} = 205$, the bottom 5\% of right singular vectors of $W_U$).
At each intervened layer, we decompose the hidden state into its dark component $h_{\mathrm{d}}$ and visible complement $h_{\perp}$, and shift weight between them:
\[
\hat h = 
\begin{cases}
(1 + \alpha)\, h_{\mathrm{d}} + (1 - \alpha)\, h_{\perp} & \text{(enhance)}, \\
(1 - \alpha)\, h_{\mathrm{d}} + (1 + \alpha)\, h_{\perp} & \text{(shrink)},
\end{cases}
\]
followed by rescaling $\hat h$ to the original norm $\|h\|_2$, where $\alpha \ge 0$ controls the intervention strength.
As elsewhere, the BOS token is left unchanged.

\paragraph{Controls.} For the control, we apply the identical operation to the matched mid-spectrum subspace used in earlier experiments (Table~\ref{tab:wikitext-middle-ranks}).

\begin{figure}[tbp]
    \centering
    \includegraphics[width=\linewidth]{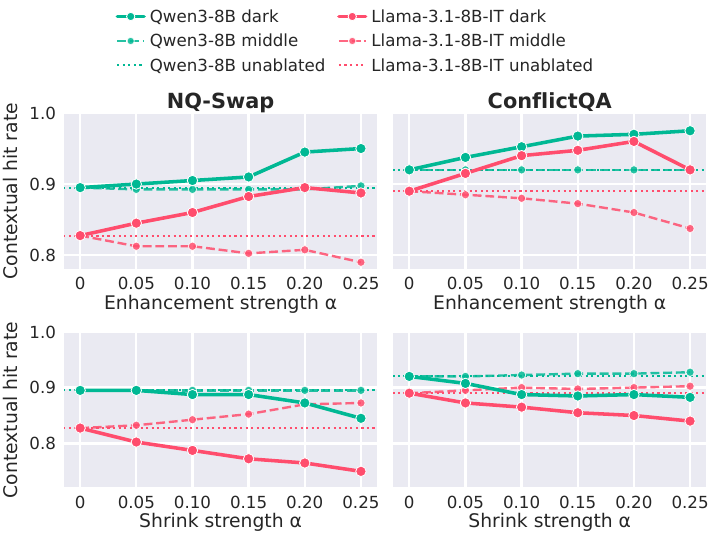}
    \caption{Contextual hit rate under residual enhancement (top) and shrinking (bottom) on layers 20--23.
    The dark component (solid) or a matched mid-spectrum component (dashed) is scaled by strength $\alpha$; dotted lines are the unablated model.
    Amplifying the dark subspace raises context use on NQ-Swap and ConflictQA, while shrinking it reduces context use; the mid-spectrum control does not show the same effect.}
    \label{fig:dm-enhance-shrink}
\end{figure}

\paragraph{Results.} Figure~\ref{fig:dm-enhance-shrink} shows that modulating dark matter bidirectionally influences context reliance on NQ-Swap and ConflictQA.
Enhancing the dark subspace raises the contextual hit rate for both models (top row).
This gain persists across $\alpha$ on Qwen3-8B, whereas on Llama-3.1-8B-Instruct it peaks around $\alpha = 0.2$ before declining, likely because large perturbations push hidden states out of distribution.
Conversely, shrinking the dark subspace suppresses context use (bottom row).
The mid-spectrum control remains near baseline on Qwen3-8B, but exhibits the inverse trend on Llama-3.1-8B-Instruct, declining under enhancement and rising under shrinking, which is consistent with shifting relative weight away from or toward the remaining directions (which contain the dark subspace).
\section{Method: DarkLens}
\label{sec:method}

Building on our empirical findings (Section~\ref{sec:empirical-study}), we introduce \textsc{DarkLens}, a lightweight faithfulness detector that balances contextual grounding against parametric reliance.
\textsc{DarkLens} pairs a parametric knowledge score of \citet{sun2025redeep} with our novel \emph{Dark Matter Ratio} ($\operatorname{DMR}$) to quantify context reliance with negligible computational overhead.

\subsection{Context--Parametric Knowledge Competition}
\label{sec:contrastive-paradigm}

Several recent works formulate faithfulness hallucination detection as an explicit competition between contextual and parametric knowledge sources~\citep{sun2025redeep, yeh2026lumina, wang2025seredeep}.
Specifically, these methods assign each generated token $t$ a hallucination score $\mathcal{H}(t)$:
\begin{equation}
\mathcal{H}(t) = \mathcal{S}_{\mathrm{param}}(t) - \mathcal{S}_{\mathrm{ctx}}(t),
\label{eq:generic-contrastive}
\end{equation}
where $\mathcal{S}_{\mathrm{param}}(t)$ quantifies the model's reliance on stored parametric weights, and $\mathcal{S}_{\mathrm{ctx}}(t)$ reflects its grounding in the retrieved passage.
When parametric reliance is elevated while contextual grounding is suppressed, the model is likely generating unsupported content, resulting in a high hallucination score.

\subsection{Dark Matter Ratio for External Context Utilization}
\label{sec:dmr}
Existing detectors in this paradigm incur substantial overhead when measuring $\mathcal{S}_{\mathrm{ctx}}(t)$, as they require materializing attention maps~\citep{sun2025redeep, wang2025seredeep}, training external probes~\citep{wang2025seredeep}, or running an additional forward pass with a counterfactual document~\citep{yeh2026lumina}.

To eliminate these bottlenecks, we build on our findings from Section~\ref{sec:empirical-study}, which show that dark components participate in context processing and that modulating their magnitude directly steers context usage.
We therefore quantify external context reliance $\mathcal{S}_{\mathrm{ctx}}(t)$ directly from the residual stream via the \emph{Dark Matter Ratio} ($\operatorname{DMR}$):
\begin{equation}
\operatorname{DMR}(h^{(\ell)}_t) = \frac{\bigl\| P_{\mathrm{dark}}\, h^{(\ell)}_t \bigr\|_2}{\bigl\| h^{(\ell)}_t \bigr\|_2},
\label{eq:dmr}
\end{equation}
where $P_{\mathrm{dark}}$ is the projection onto the dark space defined in Section~\ref{sec:background}, and $h^{(\ell)}_t$ denotes the hidden state at layer $\ell$ conditioned on prefix tokens $x_{<t}$ (i.e., the representation used immediately to predict token $t$).
Importantly, $\operatorname{DMR}$ requires only a linear projection of the hidden state, introducing negligible computational overhead during generation.

\subsection{The \textsc{DarkLens} Detector}
\label{sec:darklens-detector}

We introduce \textsc{DarkLens}, a lightweight detector that pairs our context score $\operatorname{DMR}$ with the Parametric Knowledge Score ($\mathrm{PKS}$) of \citet{sun2025redeep}.
$\mathrm{PKS}$ captures the representational shift across feed-forward blocks: letting $h^{\mathrm{mid},(\ell)}_t$ and $h^{(\ell)}_t$ denote the hidden states immediately before and after the FFN block at layer $\ell$, applying the unembedding matrix $W_U$ yields intermediate vocabulary distributions $q(h) = \mathrm{softmax}(W_U h)$.
The parametric score is the Jensen--Shannon divergence between these pre- and post-FFN distributions:
\begin{equation}
\operatorname{PKS}^{(\ell)}_t
  = \mathrm{JSD}\!\left(
       q(h^{\mathrm{mid},(\ell)}_t)
       \,\middle\|\,
       q(h^{(\ell)}_t)
     \right),
\label{eq:pks}
\end{equation}
where elevated divergence indicates that the FFN parameters have intervened strongly to steer the representation toward specific vocabulary predictions.
For brevity, we also write $\operatorname{DMR}^{(\ell)}_t = \operatorname{DMR}(h^{(\ell)}_t)$.

By substituting $\operatorname{PKS}$ for $\mathcal{S}_{\mathrm{param}}$ and $\operatorname{DMR}$ for $\mathcal{S}_{\mathrm{ctx}}$, \textsc{DarkLens} formulates the token-level hallucination score as:
\begin{equation}
\mathcal{H}(t)
  = \alpha \sum_{\ell \in \mathcal{F}} \operatorname{PKS}^{(\ell)}_t
  - \beta \, \operatorname{DMR}^{(\ell^*)}_t,
\label{eq:darklens-score}
\end{equation}
where $\mathcal{F}$ is the set of evaluated FFN layers, $\alpha, \beta > 0$ are balance weights, and the readout layer $\ell^*$ along with the dark dimension $d_{\mathrm{dark}}$ are hyperparameters selected on a small validation set.
For an entire generated response $\mathbf{r}$, the response-level hallucination score is the average token score:
\begin{equation}
\mathcal{H}(\mathbf{r}) = \frac{1}{|\mathbf{r}|} \sum_{t \in \mathbf{r}} \mathcal{H}(t).
\label{eq:response-score}
\end{equation}
Because both $\mathrm{PKS}$ and $\operatorname{DMR}$ rely solely on residual hidden states and static projection matrices ($W_U$ and $P_{\mathrm{dark}}$), \textsc{DarkLens} operates entirely within a single forward pass, remaining suitable for real-time deployment.

\section{Results}
\label{sec:detection-results}

\begin{table*}[t!]
  \centering
  \caption{AUROC scores of hallucination detection methods.
  Bold and underline indicate the best and second-best results, respectively.
  All results are averaged over 5 runs.}
  {\small \setlength{\tabcolsep}{2pt}\renewcommand{\arraystretch}{0.75}
  \begin{tabular}{l|ccc|ccc}\toprule
   & \multicolumn{3}{c}{MS MARCO} & \multicolumn{3}{c}{CNN/DM} \\
   & Mistral-7B & Llama-2-7B & Llama-2-13B & Mistral-7B & Llama-2-7B & Llama-2-13B \\
  \midrule
  Ours
    & \textbf{0.77 \scriptsize{$\pm$ 0.02}}
    & \textbf{0.71 \scriptsize{$\pm$ 0.04}}
    & \textbf{0.72 \scriptsize{$\pm$ 0.04}}
    & \textbf{0.66 \scriptsize{$\pm$ 0.07}}
    & 0.58 \scriptsize{$\pm$ 0.03}
    & \textbf{0.59 \scriptsize{$\pm$ 0.05}} \\
  LLM Check
    & 0.49 \scriptsize{$\pm$ 0.05}
    & 0.47 \scriptsize{$\pm$ 0.07}
    & 0.51 \scriptsize{$\pm$ 0.05}
    & 0.54 \scriptsize{$\pm$ 0.04}
    & 0.50 \scriptsize{$\pm$ 0.06}
    & 0.54 \scriptsize{$\pm$ 0.03} \\
  Focus
    & 0.67 \scriptsize{$\pm$ 0.06}
    & 0.63 \scriptsize{$\pm$ 0.03}
    & 0.61 \scriptsize{$\pm$ 0.03}
    & 0.62 \scriptsize{$\pm$ 0.04}
    & 0.51 \scriptsize{$\pm$ 0.09}
    & 0.50 \scriptsize{$\pm$ 0.06} \\
  LN-Entropy
    & 0.65 \scriptsize{$\pm$ 0.06}
    & 0.63 \scriptsize{$\pm$ 0.03}
    & 0.62 \scriptsize{$\pm$ 0.04}
    & \underline{0.65 \scriptsize{$\pm$ 0.04}}
    & \textbf{0.63 \scriptsize{$\pm$ 0.09}}
    & 0.56 \scriptsize{$\pm$ 0.04} \\
  LSC
    & 0.68 \scriptsize{$\pm$ 0.04}
    & 0.57 \scriptsize{$\pm$ 0.03}
    & 0.57 \scriptsize{$\pm$ 0.05}
    & 0.64 \scriptsize{$\pm$ 0.04}
    & \underline{0.62 \scriptsize{$\pm$ 0.09}}
    & 0.56 \scriptsize{$\pm$ 0.05} \\
  LUMINA
    & 0.66 \scriptsize{$\pm$ 0.05}
    & \textbf{0.71 \scriptsize{$\pm$ 0.04}}
    & \underline{0.70 \scriptsize{$\pm$ 0.04}}
    & 0.64 \scriptsize{$\pm$ 0.07}
    & 0.54 \scriptsize{$\pm$ 0.06}
    & 0.54 \scriptsize{$\pm$ 0.04} \\
  MaxProb
    & \underline{0.75 \scriptsize{$\pm$ 0.04}}
    & 0.69 \scriptsize{$\pm$ 0.03}
    & 0.67 \scriptsize{$\pm$ 0.04}
    & 0.61 \scriptsize{$\pm$ 0.07}
    & 0.51 \scriptsize{$\pm$ 0.07}
    & 0.53 \scriptsize{$\pm$ 0.08} \\
  Perplexity
    & 0.64 \scriptsize{$\pm$ 0.05}
    & 0.48 \scriptsize{$\pm$ 0.03}
    & 0.46 \scriptsize{$\pm$ 0.03}
    & 0.49 \scriptsize{$\pm$ 0.08}
    & 0.52 \scriptsize{$\pm$ 0.04}
    & 0.48 \scriptsize{$\pm$ 0.04} \\
  RAUQ
    & 0.66 \scriptsize{$\pm$ 0.04}
    & 0.60 \scriptsize{$\pm$ 0.02}
    & 0.59 \scriptsize{$\pm$ 0.04}
    & 0.58 \scriptsize{$\pm$ 0.06}
    & 0.50 \scriptsize{$\pm$ 0.10}
    & 0.51 \scriptsize{$\pm$ 0.05} \\
  ReDeEP
    & \underline{0.72 \scriptsize{$\pm$ 0.03}}
    & 0.63 \scriptsize{$\pm$ 0.05}
    & 0.66 \scriptsize{$\pm$ 0.03}
    & 0.63 \scriptsize{$\pm$ 0.08}
    & 0.59 \scriptsize{$\pm$ 0.03}
    & \underline{0.57 \scriptsize{$\pm$ 0.08}} \\
  SAR
    & 0.67 \scriptsize{$\pm$ 0.04}
    & 0.55 \scriptsize{$\pm$ 0.02}
    & 0.48 \scriptsize{$\pm$ 0.03}
    & 0.57 \scriptsize{$\pm$ 0.04}
    & 0.48 \scriptsize{$\pm$ 0.10}
    & 0.52 \scriptsize{$\pm$ 0.05} \\
  Semantic Density
    & 0.55 \scriptsize{$\pm$ 0.02}
    & 0.55 \scriptsize{$\pm$ 0.05}
    & 0.53 \scriptsize{$\pm$ 0.02}
    & 0.56 \scriptsize{$\pm$ 0.08}
    & 0.52 \scriptsize{$\pm$ 0.04}
    & \underline{0.57 \scriptsize{$\pm$ 0.07}} \\
  Semantic Entropy
    & 0.60 \scriptsize{$\pm$ 0.05}
    & 0.53 \scriptsize{$\pm$ 0.04}
    & 0.62 \scriptsize{$\pm$ 0.04}
    & 0.56 \scriptsize{$\pm$ 0.06}
    & 0.57 \scriptsize{$\pm$ 0.06}
    & 0.52 \scriptsize{$\pm$ 0.03} \\
  \midrule
  & \multicolumn{6}{c}{XSum} \\
  \midrule
  & Mistral-7B & Llama-2-7B & \multicolumn{1}{c}{Llama-2-13B} & Llama-3.1-8B & Qwen3-8B & gemma-3-4b \\
\midrule
  Ours
  & \underline{0.66 \scriptsize{$\pm$ 0.04}}
  & \textbf{0.65 \scriptsize{$\pm$ 0.06}}
  & \multicolumn{1}{c}{\textbf{0.66 \scriptsize{$\pm$ 0.02}}}
  & \textbf{0.65 \scriptsize{$\pm$ 0.02}}
  & \textbf{0.71 \scriptsize{$\pm$ 0.05}}
  & \textbf{0.71 \scriptsize{$\pm$ 0.07}} \\
LLM Check
  & 0.59 \scriptsize{$\pm$ 0.03}
  & 0.50 \scriptsize{$\pm$ 0.08}
  & \multicolumn{1}{c}{0.56 \scriptsize{$\pm$ 0.04}}
  & 0.51 \scriptsize{$\pm$ 0.05}
  & 0.46 \scriptsize{$\pm$ 0.06}
  & 0.50 \scriptsize{$\pm$ 0.02} \\
Focus
  & 0.63 \scriptsize{$\pm$ 0.08}
  & \textbf{0.65 \scriptsize{$\pm$ 0.06}}
  & \multicolumn{1}{c}{0.58 \scriptsize{$\pm$ 0.04}}
  & 0.62 \scriptsize{$\pm$ 0.05}
  & 0.61 \scriptsize{$\pm$ 0.03}
  & 0.61 \scriptsize{$\pm$ 0.04} \\
LN-Entropy
  & \textbf{0.69 \scriptsize{$\pm$ 0.03}}
  & 0.61 \scriptsize{$\pm$ 0.02}
  & \multicolumn{1}{c}{0.62 \scriptsize{$\pm$ 0.04}}
  & \underline{0.64 \scriptsize{$\pm$ 0.03}}
  & 0.65 \scriptsize{$\pm$ 0.04}
  & 0.57 \scriptsize{$\pm$ 0.06} \\
LSC
  & 0.64 \scriptsize{$\pm$ 0.05}
  & 0.57 \scriptsize{$\pm$ 0.05}
  & \multicolumn{1}{c}{0.58 \scriptsize{$\pm$ 0.04}}
  & 0.59 \scriptsize{$\pm$ 0.08}
  & 0.62 \scriptsize{$\pm$ 0.04}
  & 0.53 \scriptsize{$\pm$ 0.07} \\
LUMINA
  & 0.62 \scriptsize{$\pm$ 0.03}
  & 0.58 \scriptsize{$\pm$ 0.06}
  & \multicolumn{1}{c}{0.59 \scriptsize{$\pm$ 0.04}}
  & 0.56 \scriptsize{$\pm$ 0.05}
  & 0.61 \scriptsize{$\pm$ 0.05}
  & 0.54 \scriptsize{$\pm$ 0.05} \\
MaxProb
  & 0.60 \scriptsize{$\pm$ 0.07}
  & 0.58 \scriptsize{$\pm$ 0.05}
  & \multicolumn{1}{c}{0.58 \scriptsize{$\pm$ 0.04}}
  & 0.62 \scriptsize{$\pm$ 0.04}
  & 0.63 \scriptsize{$\pm$ 0.04}
  & 0.53 \scriptsize{$\pm$ 0.05} \\
Perplexity
  & 0.58 \scriptsize{$\pm$ 0.05}
  & 0.56 \scriptsize{$\pm$ 0.05}
  & \multicolumn{1}{c}{0.53 \scriptsize{$\pm$ 0.04}}
  & 0.61 \scriptsize{$\pm$ 0.05}
  & 0.63 \scriptsize{$\pm$ 0.03}
  & 0.47 \scriptsize{$\pm$ 0.07} \\
RAUQ
  & \underline{0.66 \scriptsize{$\pm$ 0.03}}
  & 0.59 \scriptsize{$\pm$ 0.04}
  & \multicolumn{1}{c}{0.59 \scriptsize{$\pm$ 0.04}}
  & \underline{0.63 \scriptsize{$\pm$ 0.04}}
  & 0.63 \scriptsize{$\pm$ 0.03}
  & 0.56 \scriptsize{$\pm$ 0.08} \\
ReDeEP
  & 0.62 \scriptsize{$\pm$ 0.04}
  & 0.61 \scriptsize{$\pm$ 0.04}
  & \multicolumn{1}{c}{0.60 \scriptsize{$\pm$ 0.04}}
  & \underline{0.63 \scriptsize{$\pm$ 0.02}}
  & \underline{0.69 \scriptsize{$\pm$ 0.05}}
  & \underline{0.62 \scriptsize{$\pm$ 0.06}} \\
SAR
  & \underline{0.66 \scriptsize{$\pm$ 0.03}}
  & 0.60 \scriptsize{$\pm$ 0.03}
  & \multicolumn{1}{c}{0.58 \scriptsize{$\pm$ 0.05}}
  & \underline{0.63 \scriptsize{$\pm$ 0.04}}
  & 0.64 \scriptsize{$\pm$ 0.03}
  & 0.50 \scriptsize{$\pm$ 0.05} \\
Semantic Density
  & 0.63 \scriptsize{$\pm$ 0.03}
  & 0.63 \scriptsize{$\pm$ 0.06}
  & \multicolumn{1}{c}{0.59 \scriptsize{$\pm$ 0.05}}
  & 0.55 \scriptsize{$\pm$ 0.06}
  & 0.60 \scriptsize{$\pm$ 0.07}
  & 0.58 \scriptsize{$\pm$ 0.06} \\
Semantic Entropy
  & 0.64 \scriptsize{$\pm$ 0.05}
  & 0.63 \scriptsize{$\pm$ 0.03}
  & \multicolumn{1}{c}{\underline{0.62 \scriptsize{$\pm$ 0.02}}}
  & 0.61 \scriptsize{$\pm$ 0.05}
  & 0.67 \scriptsize{$\pm$ 0.04}
  & \underline{0.62 \scriptsize{$\pm$ 0.06}} \\
\bottomrule
  \end{tabular}}
  \label{tab:ragtruth-auroc}
  \end{table*}

\paragraph{Datasets and models.} We evaluate on three question answering and summarization tasks: long-form QA (MS MARCO) and summarization (CNN/DM) from RAGTruth~\cite{niu2024ragtruth}, and extreme summarization (XSum~\cite{narayan2018don}).

We employ six open-source LLMs that provide access to their internal states: Mistral-7B-Instruct-v0.1~\citep{jiang2023mistral7b}, LLaMA-2-7B-chat and LLaMA-2-13B-chat~\citep{touvron2023llama}, LLaMA-3.1-8B-Instruct~\citep{grattafiori2024llama}, Qwen3-8B~\citep{qwen3technicalreport}, and gemma-3-4b-it~\citep{gemma_2025}. Because RAGTruth provides annotations only for Mistral-7B and the LLaMA-2 models, evaluation on MS MARCO and CNN/DM is restricted to them.

\paragraph{Baselines.} We compare our method against RAG-specific detectors and against uncertainty and internals baselines. The closest methods are ReDeEP~\citep{sun2025redeep} and LUMINA~\citep{yeh2026lumina}. The information-based baselines are LN-Entropy~\citep{malinin2020uncertainty}, Focus~\citep{zhang2023enhancing}, Lowest Span Confidence (LSC)~\citep{qiao2026lowest}, LLM-Check~\citep{sriramanan2024llm}, MaxProb~\citep{fadeeva2024fact}, Perplexity~\citep{fomicheva2020unsupervised}, and RAUQ~\citep{vazhentsev2025uncertainty}. Sampling-based methods are SAR~\citep{duan2024shifting}, Semantic Density~\citep{qiu2024semantic}, and Semantic Entropy~\citep{kuhn2023semantic}.

\paragraph{Main results.}
Table~\ref{tab:ragtruth-auroc} reports response-level AUROC against uncertainty, sampling, and internals baselines. We report the mean and standard deviation across 5-fold cross-validation.
On MS MARCO, \textsc{DarkLens} attains the top score across all three models, outperforming ReDeEP by 5--8 points.
On CNN/DM, it remains strongest on Mistral-7B and Llama-2-13B, while LN-Entropy and LSC lead on Llama-2-7B.
On XSum, \textsc{DarkLens} achieves the best result on five of six models and places second on Mistral-7B, posting its highest scores on Qwen3-8B and gemma-3-4b.
Crucially, \textsc{DarkLens} consistently surpasses both ReDeEP and LUMINA across all six models on XSum.
Overall, \textsc{DarkLens} performs competitively against both general-purpose and RAG-specific baselines, matching or exceeding the strongest of them in most evaluated settings.

\begin{figure}[t]
  \centering
  \includegraphics[width=0.9\linewidth]{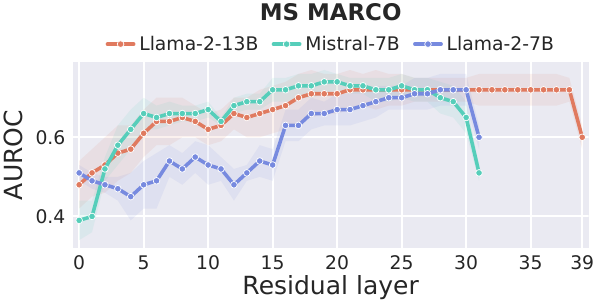}
  \caption{Response-level AUROC of $-\operatorname{DMR}$ as a function of the residual readout layer on MS MARCO.}
  \label{fig:auroc-vs-layer}
\end{figure}

\paragraph{Complementarity of PKS and DMR.}
Table~\ref{tab:ablation} ablates the individual contributions of $\mathrm{PKS}$ and $\operatorname{DMR}$.
Note that because greater context reliance corresponds to lower hallucination risk, $-\operatorname{DMR}$ is used as the standalone detector score in this ablation, as well as in Figure~\ref{fig:auroc-vs-layer} and Table~\ref{tab:subspace-ablation}. 
The results show that both terms are important for \textsc{DarkLens}: while $\operatorname{DMR}$ alone carries most of the discriminative power in the majority of configurations, pairing it with $\mathrm{PKS}$ yields clear improvements in specific cases, notably on XSum for Llama-3.1-8B (0.61 $\to$ 0.65) and gemma-3-4b (0.69 $\to$ 0.71).

\begin{figure}[t]
  \centering
  \includegraphics[width=\linewidth]{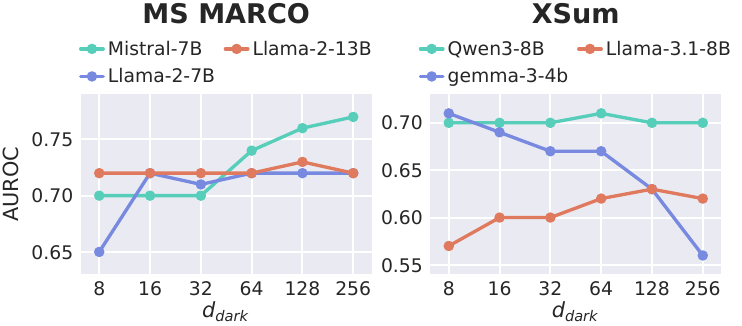}
  \caption{Response-level AUROC of \textsc{DarkLens} as a function of the dark subspace rank ($d_{\mathrm{dark}}$) on MS MARCO (left) and XSum (right).}
  \label{fig:auroc-vs-null-dim}
\end{figure}

\paragraph{Sensitivity to dark dimension.}
Figure~\ref{fig:auroc-vs-null-dim} reports \textsc{DarkLens} detection AUROC across varying ranks $d_{\mathrm{dark}}$.
We observe that while extremely small ranks ($d_{\mathrm{dark}} = 8$) may fail to reliably isolate the context signal and excessively large ranks ($d_{\mathrm{dark}} = 256$ on gemma-3-4b, representing $10\%$ of its dimension) may cause degradation by encroaching into the vocabulary-visible subspace, performance is largely stable between these extremes.
In practice, selecting $d_{\mathrm{dark}}$ within approximately 1.5--3\% of $d_{\mathrm{model}}$ serves as a safe rule of thumb when tuning on validation data is not desired.

\paragraph{Sensitivity to readout layer.}
Figure~\ref{fig:auroc-vs-layer} reports detection AUROC as a function of the residual layer used for $\operatorname{DMR}$ readout on MS MARCO (in this experiment, $d_{\textrm{dark}} = 64$).
Performance generally improves with depth, remaining stable across mid-to-late layers before dropping near the final layer.
This drop matches what \citet{cancedda2024spectral} observed for the BOS token: in the final layers, representations leave the dark space as the model prepares its final token prediction.
When tuning on a validation split is not feasible, selecting $\ell^*$ at roughly 80\% of the model depth serves as a robust default.

\paragraph{Specificity of the dark space.}
We evaluate whether detection performance reflects the dark space properties specifically or merely low-dimensional projection in general.
In Table~\ref{tab:subspace-ablation}, replacing the dark subspace with equal-dimension ($d=64$) mid-spectrum or random orthogonal subspaces degrades detection to near-chance levels across benchmarks, supporting the hypothesis that context-reliance information is primarily carried by directions with low $W_U$-composition.

\begin{table}[t]
\centering
\caption{Ablation of DarkLens components (AUROC score, mean $\pm$ std over 5 runs) on MS MARCO and XSum. PKS: parametric knowledge score only;  DMR: standalone negative dark matter ratio ($-\operatorname{DMR}$); Both: combined DarkLens detector.}
\label{tab:ablation}
{\small \setlength{\tabcolsep}{3pt}\renewcommand{\arraystretch}{0.85}
\begin{tabular}{lccc}
\toprule
& \multicolumn{3}{c}{MS MARCO} \\
\cmidrule(lr){2-4}
Ablation & Mistral-7B & Llama-2-7B & Llama-2-13B \\
\midrule
PKS  & 0.74 \scriptsize{$\pm$ 0.01} & 0.63 \scriptsize{$\pm$ 0.04} & 0.70 \scriptsize{$\pm$ 0.02} \\
DMR  & 0.77 \scriptsize{$\pm$ 0.02} & 0.72 \scriptsize{$\pm$ 0.03} & 0.72 \scriptsize{$\pm$ 0.04} \\
Both & 0.77 \scriptsize{$\pm$ 0.02} & 0.71 \scriptsize{$\pm$ 0.04} & 0.72 \scriptsize{$\pm$ 0.04} \\
\midrule
& \multicolumn{3}{c}{XSum} \\
\cmidrule(lr){2-4}
Ablation & Qwen3-8B & gemma-3-4b & Llama-3.1-8B \\
\midrule
PKS  & 0.69 \scriptsize{$\pm$ 0.04} & 0.64 \scriptsize{$\pm$ 0.06} & 0.65 \scriptsize{$\pm$ 0.03} \\
DMR  & 0.71 \scriptsize{$\pm$ 0.05} & 0.69 \scriptsize{$\pm$ 0.07} & 0.61 \scriptsize{$\pm$ 0.02} \\
Both & 0.71 \scriptsize{$\pm$ 0.05} & 0.71 \scriptsize{$\pm$ 0.05} & 0.65 \scriptsize{$\pm$ 0.02} \\
\bottomrule
\end{tabular}}
\end{table}

\begin{table}[t]
\centering
\caption{Subspace ablation for $-\operatorname{DMR}$ (AUROC, mean $\pm$ std over 5 runs).
Dark: bottom right singular vectors of $W_U$; middle: matched mid-spectrum subspace; random: random subspace of equal rank.}
\label{tab:subspace-ablation}
{\small \setlength{\tabcolsep}{2pt}\renewcommand{\arraystretch}{0.8}
\begin{tabular}{lcccc}
\toprule
& \multicolumn{2}{c}{MS MARCO} & \multicolumn{2}{c}{XSum} \\
\cmidrule(lr){2-3} \cmidrule(lr){4-5}
Subspace & Mistral-7B & Llama-2-7B & Qwen3-8B & gemma-3-4b \\
\midrule
Dark   & 0.74 \scriptsize{$\pm$ 0.03} & 0.72 \scriptsize{$\pm$ 0.03} & 0.71 \scriptsize{$\pm$ 0.05} & 0.67 \scriptsize{$\pm$ 0.05} \\
Middle & 0.57 \scriptsize{$\pm$ 0.05} & 0.50 \scriptsize{$\pm$ 0.05} & 0.56 \scriptsize{$\pm$ 0.08} & 0.56 \scriptsize{$\pm$ 0.05} \\
Random & 0.51 \scriptsize{$\pm$ 0.04} & 0.52 \scriptsize{$\pm$ 0.05} & 0.47 \scriptsize{$\pm$ 0.09} & 0.57 \scriptsize{$\pm$ 0.05} \\
\bottomrule
\end{tabular}}
\end{table}

\section{Related Work}
\label{sec:related-work}

\paragraph{Geometry of the unembedding matrix.}
Projecting intermediate hidden states through the unembedding matrix~---~via the logit lens~\citep{logitlens} and tuned lens~\citep{belrose2023eliciting}~---~has become a standard technique for analyzing representations in both LLMs and VLMs~\cite{arad-etal-2025-saes,wang2026same,wang2026truth}.

By construction, however, these tools largely discard the unembedding dark space (see Section~\ref{sec:background}), leaving this subspace sparsely studied. Prior work associated this weakly composing ``dark space'' primarily with BOS attention sinks~\citep{cancedda2024spectral} and confidence calibration~\citep{stolfo2024confidence,joshi2025calibration}. To the best of our knowledge, the only prior link of the dark space to context reliance is by \citet{tighidet-etal-2025-context}, who found that a few final-layer dark-space ``entropy neurons''~\citep{stolfo2024confidence} suppress context copying; in contrast, we study intermediate dark representations, showing that they actively drive context utilization.

\paragraph{Internal-state faithfulness detection.}
While internal-state detectors avoid external critics or repeated sampling~\cite{chuang2024lookback, binkowski2025hallucination, bazarova-etal-2026-hallucination}, most rely on learned probes or heuristics that treat internal states as black boxes, overlooking why the model hallucinated~\cite{binkowski2026attention, qi2026detecting}.
Recent work addresses this by explicitly framing faithfulness as a competition between stored parametric memory and retrieved context~\cite{sun2025redeep,yeh2026lumina,wang2025seredeep}.
However, the way these methods measure context reliance introduces non-negligible overhead:
ReDeEP~\citep{sun2025redeep} aggregates context representations attended by copy heads, requiring explicit attention map materialization rather than relying solely on residual states;
LUMINA~\citep{yeh2026lumina} measures distribution shifts under random document substitutions, requiring an additional forward pass;
and SeReDeEP~\citep{wang2025seredeep} replaces ReDeEP's geometric readout with semantic-entropy probes, adding auxiliary supervision without removing the attention bottleneck.
Instead of relying on attention weights or input perturbations, \textsc{DarkLens} reads context utilization directly from residual stream geometry via the dark matter ratio ($\operatorname{DMR}$), requiring only a static projection at inference time.

\section{Conclusion}
In this work, we examined the functional role of the unembedding dark space~---~the subspace of residual directions that compose weakly with the unembedding matrix $W_U$ and minimally affect vocabulary logits~---~in autoregressive language models. Contrary to prior hypotheses that this subspace primarily enables BOS attention sinks, our empirical findings suggest that dark components at subsequent positions actively mediate context retrieval and processing. Building on this insight, we introduced the Dark Matter Ratio ($\operatorname{DMR}$), which measures the fraction of hidden-state norm residing in the dark space to quantify context reliance directly from residual stream geometry. Combining $\operatorname{DMR}$ with a parametric knowledge score, our resulting detector \textsc{DarkLens} reliably identifies faithfulness hallucinations across QA and summarization benchmarks within a single forward pass, matching or outperforming a broad range of baselines without the overhead of extracting attention maps or training auxiliary models.

\section*{Limitations}

\begin{itemize}
    \item \textbf{Faithfulness vs.\ Factuality:} \textsc{DarkLens} measures faithfulness rather than factuality. If a model is given inaccurate context and faithfully repeats it, \textsc{DarkLens} will not flag the output as an error.

\item \textbf{White-Box Access:} Although \textsc{DarkLens} avoids materializing attention maps, it still requires access to residual stream activations, making it inapplicable to closed-source, black-box models.

\item \textbf{Underlying Mechanism:} The precise circuit-level mechanism by which the dark space mediates context retrieval remains incompletely understood. We hypothesize that the model may route information through dark directions to modulate attention heads, which we leave for future investigation.
\end{itemize}



\bibliography{custom}

\newpage
\appendix

\section{Synthetic induction: experimental details}
\label{sec:app-synth-induction}

\paragraph{NLL-matched mid-spectrum ranks.} Table~\ref{tab:wikitext-middle-ranks} lists the mid-spectrum subspaces used for the matched-NLL controls in Section~\ref{sec:empirical-study}.
Across all windows, the dark subspace spans the bottom 5\% of the right singular vectors of $W_U$ (rank $205$, vectors $3891$--$4095$, with $d_{\text{model}} = 4096$).
We evaluate negative log-likelihood on $128$ sequences of $256$ tokens from the WikiText-2 test split, and select the centered middle band of smallest rank whose mean NLL falls within $0.01$ of the corresponding dark-swap intervention, ensuring that the dark components themselves are never included in the middle subspace.
  
\begin{table}[t]
  \centering
  \small
  \begin{tabular}{lrrrr}
  \toprule
  & \multicolumn{2}{c}{Llama-3.1-8B-It} & \multicolumn{2}{c}{Qwen3-8B} \\
  \cmidrule(lr){2-3} \cmidrule(lr){4-5}
  Layers & Rank & Sing. vectors & Rank & Sing. vectors \\
  \midrule
  0--3   & 1636 & 1230--2865 & 1867 & 1114--2980 \\
  4--7   &  904 & 1596--2499 &  782 & 1657--2438 \\
  8--11  & 1174 & 1461--2634 &  205 & 1945--2149 \\
  12--15 &  954 & 1571--2524 & 2706 &  695--3400 \\
  16--19 &  760 & 1668--2427 & 3690 &  201--3890 \\
  20--23 &  883 & 1606--2488 &  205 & 1945--2149 \\
  24--27 & 1426 & 1335--2760 & 1005 & 1545--2549 \\
  28--31 & 3891 &    0--3890 &  500 & 1798--2297 \\
  32--35 &   -- &         -- & 3280 &  408--3687 \\
  \bottomrule
  \end{tabular}
  \caption{Matched mid-spectrum ranks on WikiText-2. For each layer window, swapping within the indicated middle subspace reproduces the NLL increase of swapping the rank-205 dark subspace (bottom 5\% of right singular vectors of $W_U$) within $\pm 0.01$.}
  \label{tab:wikitext-middle-ranks}
  \end{table}

\begin{figure}[t]
  \centering
  \includegraphics[width=\linewidth]{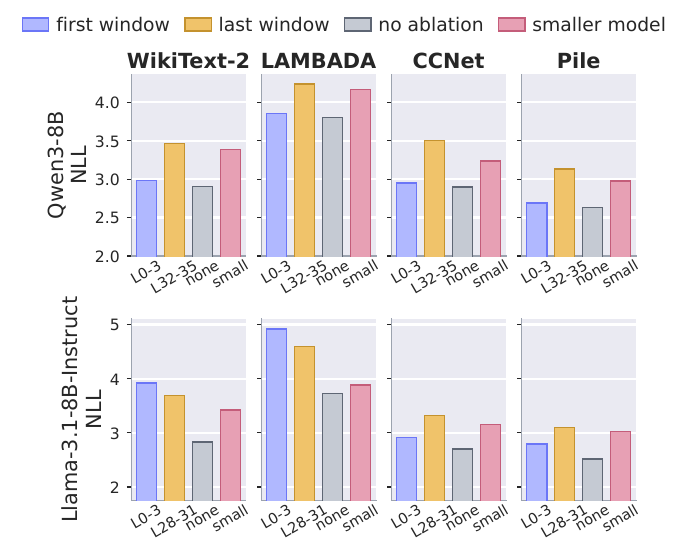}
  \caption{Language-modeling NLL under dark residual swaps on the first and last four-layer windows.
  Blue: first window (layers 0--3); orange: last window (layers 32--35 for Qwen3-8B, 28--31 for Llama-3.1-8B-Instruct); grey: unablated model; pink: smaller reference model (Qwen3-0.6B and Llama-3.2-1B-Instruct).
  Boundary interventions raise NLL above the unablated model and, on most corpora, to or above the smaller model.}
  \label{fig:last-layers-nll}
\end{figure}

\paragraph{NLL damage at the first and last windows.} 
Figure~\ref{fig:last-layers-nll} reports language-modeling NLL under dark swaps applied exclusively to the first and last four-layer windows.
For both models, intervening in the final window degrades NLL substantially.
This is expected: although singular values in the dark subspace are small, they are non-zero.
In the final layer, dark components account for a substantial fraction of the hidden-state norm (58\% for Llama-3.1-8B-Instruct and 77\% for Qwen3-8B; measured on 100 sequences of length 256 from WikiText-2), so their residual contribution to the final logits remains non-negligible (25\% and 11\% of total logit norm, respectively).
Perturbing dark components in the final layers thus directly corrupts output logits.
Conversely, intervening in the first window impairs Llama but leaves Qwen largely unaffected, indicating that early-layer sensitivity to dark-subspace perturbations is architecture- or training-dependent rather than universal.
Because these boundary interventions induce severe general degradation that confounds context-specific effects, we exclude them from the main experiments in Section~\ref{sec:empirical-study}.

\begin{figure}[t]
  \centering
  \includegraphics[width=\linewidth]{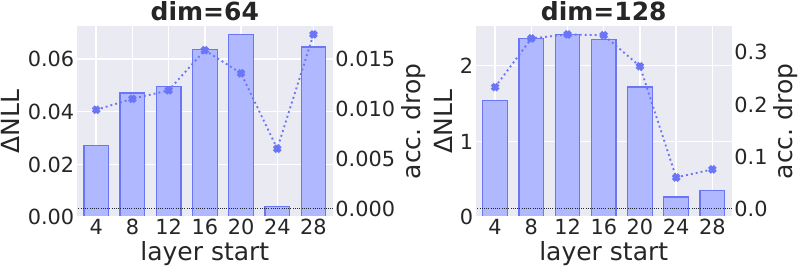}
  \caption{Synthetic induction dark swaps on Qwen3-8B with reduced dark subspace ranks ($d_{\mathrm{dark}} = 64$ and $128$).
  Bars show $\Delta$NLL on the repeated half against the start layer of each four-layer window; the dotted line is the average accuracy drop.}
  \label{fig:induction-qwen-dark-rank}
\end{figure}

\paragraph{Sensitivity to dark subspace rank.}
Figure~\ref{fig:induction-qwen-dark-rank} repeats the Qwen3-8B synthetic induction dark-swap experiment of Section~\ref{sec:synthetic-induction} with reduced ranks $d_{\mathrm{dark}} \in \{64, 128\}$ (instead of the main-text rank $205$).
Dark swaps continue to impair copying on the repeated half at both ranks, with the effect strengthening as $d_{\mathrm{dark}}$ increases.

\section{Contextual vs Parametric: experimental details}
\label{sec:app-ctx-vs-param}

\paragraph{NLL-matched mid-spectrum ranks.} Table~\ref{tab:wikitext-middle-perm-nll-match} lists the mid-spectrum subspaces used for the matched-NLL permutation controls in Section~\ref{sec:empirical-study}.
As in the synthetic induction experiments, the dark subspace spans the bottom 5\% of the right singular vectors of $W_U$ (rank $205$, vectors $3891$--$4095$, with $d_{\text{model}} = 4096$).
The middle subspace is selected using the same procedure as in Appendix~\ref{sec:app-synth-induction}.

\begin{table}[t]
  \centering
  \small
  \caption{Matched mid-spectrum ranks on WikiText-2 (in-sample permutation). For each layer window, permuting within the indicated middle subspace reproduces the NLL increase of permuting the rank-205 dark subspace (bottom 5\% of right singular vectors of $W_U$) within $\pm 0.01$.}
  \begin{tabular}{lcccc}
  \toprule
   & \multicolumn{2}{c}{Llama-3.1-8B-It} & \multicolumn{2}{c}{Qwen3-8B} \\
  \cmidrule(lr){2-3}\cmidrule(lr){4-5}
  Layers & Rank & Sing.\ vectors & Rank & Sing.\ vectors \\
  \midrule
  0–3 & 1112 & 1492–2603 & 2161 & 967–3127 \\
  4–7 & 968 & 1564–2531 & 2107 & 994–3100 \\
  8–11 & 1200 & 1448–2647 & 3056 & 520–3575 \\
  12–15 & 1130 & 1483–2612 & 3270 & 413–3682 \\
  16–19 & 974 & 1561–2534 & 3085 & 505–3589 \\
  20–23 & 1046 & 1525–2570 & 2648 & 724–3371 \\
  24–27 & 1488 & 1304–2791 & 2038 & 1029–3066 \\
  28–31 & 2314 & 891–3204 & 1668 & 1214–2881 \\
  32–35 & – & – & 2673 & 711–3383 \\
  \bottomrule
  \end{tabular}
  \label{tab:wikitext-middle-perm-nll-match}
  \end{table}

\paragraph{Interventions across layer windows.}\label{sec:app-results}
Table~\ref{tab:subspace-hit-rates-appendix} extends the evaluation in Table~\ref{tab:subspace-hit-rates} to layer windows 16--19 and 24--27.
The observed trends align closely with the main findings in Section~\ref{sec:empirical-study}.
For Qwen3-8B, dark subspace permutations substantially reduce context-aligned preference on NQ-Swap and ConflictQA---especially in layers 16--19---while leaving parametric factual accuracy largely intact. In contrast, the mid-spectrum control in layers 16--19 degrades both parametric and contextual performance, showing no selectivity for knowledge source.
For Llama-3.1-8B-Instruct, parametric recall is unaffected by either intervention, while the dark permutation consistently lowers contextual preference compared to the mid-spectrum control.
Together, these results indicate that the selective impairment of context utilization under dark interventions holds across multiple layer windows.

\begin{table*}[tbp]
  \centering
  \small
  \setlength{\tabcolsep}{4pt}
  \begin{tabular}{lllccccc}
  \toprule
  Model & Type & Dataset & No ablation & \multicolumn{2}{c}{L16--19} & \multicolumn{2}{c}{L24--27} \\
  \cmidrule(lr){5-6} \cmidrule(lr){7-8}
  & & & & Middle & Dark & Middle & Dark \\
  \midrule
  \multirow{5}{*}{Qwen3-8B}
    & \multirow{2}{*}{Contextual} & NQ-Swap    & 0.88 & 0.61 & 0.30 & 0.92 & 0.62 \\
    &                             & ConflictQA & 0.98 & 0.98 & 0.69 & 0.99 & 0.78 \\
    & \multirow{3}{*}{Parametric} & Cities     & 1.0  & 0.63 & 0.99 & 1.0  & 1.0  \\
    &                             & Sports     & 0.95 & 0.55 & 0.92 & 0.93 & 0.95 \\
    &                             & Celebrity  & 0.96 & 0.62 & 0.93 & 0.96 & 0.96 \\
  \addlinespace
  \multirow{5}{*}{Llama-3.1-8B}
    & \multirow{2}{*}{Contextual} & NQ-Swap    & 0.83 & 0.88 & 0.54 & 0.86 & 0.73 \\
    &                             & ConflictQA & 0.96 & 0.96 & 0.89 & 0.95 & 0.93 \\
    & \multirow{3}{*}{Parametric} & Cities     & 1.0  & 1.0  & 1.0  & 0.99 & 1.0  \\
    &                             & Sports     & 0.95 & 0.94 & 0.93 & 0.95 & 0.94 \\
    &                             & Celebrity  & 0.96 & 0.95 & 0.96 & 0.95 & 0.95 \\
  \bottomrule
  \end{tabular}
  \caption{Same comparison as Table~\ref{tab:subspace-hit-rates} for layers 16--19 and 24--27. Dark subspace dimension 205.}
  \label{tab:subspace-hit-rates-appendix}
  \end{table*}

\paragraph{Sensitivity to dark subspace rank.}
Tables~\ref{tab:qwen-dark-hit-nd32}--\ref{tab:qwen-dark-hit-nd128} present supplementary results for the experiment in Section~\ref{sec:ctx-vs-param}, reporting target hit rates on contextual and parametric datasets for Qwen3-8B across dark subspace ranks $d_{\mathrm{dark}} \in \{32, 64, 128\}$.
Although the impact diminishes as $d_{\mathrm{dark}}$ decreases, it remains substantial even at smaller dimensions: on NQ-Swap with $d_{\mathrm{dark}} = 64$, the target hit rate drops by 13\% in relative terms.
In contrast, performance on parametric datasets remains completely unchanged regardless of the dark subspace dimension.

\begin{table}
\centering
\small
\caption{Qwen3-8B contextual/factual hit rate under dark residual permutation (dark rank 32). Hit = target continuation has lower mean NLL than the alternative.}
\label{tab:qwen-dark-hit-nd32}
\begin{tabular}{lccc}
\toprule
Dataset & no ablation & L16–19 & L20–23 \\
\midrule
NQ-Swap & 0.88 & 0.82 & 0.76 \\
ConflictQA & 0.98 & 0.98 & 0.98 \\
Cities & 1.00 & 0.99 & 1.00 \\
Sports & 0.95 & 0.94 & 0.96 \\
Celebrity & 0.96 & 0.96 & 0.96 \\
\bottomrule
\end{tabular}
\end{table}

\begin{table}
\centering
\small
\caption{Qwen3-8B contextual/factual hit rate under dark residual permutation (dark rank 64). Hit = target continuation has lower mean NLL than the alternative.}
\label{tab:qwen-dark-hit-nd64}
\begin{tabular}{lccc}
\toprule
Dataset & no ablation & L16–19 & L20–23 \\
\midrule
NQ-Swap & 0.88 & 0.77 & 0.63 \\
ConflictQA & 0.98 & 0.98 & 0.97 \\
Cities & 1.00 & 1.00 & 1.00 \\
Sports & 0.95 & 0.94 & 0.94 \\
Celebrity & 0.96 & 0.94 & 0.96 \\
\bottomrule
\end{tabular}
\end{table}

\begin{table}
  \centering
\small
\caption{Qwen3-8B contextual/factual hit rate under dark residual permutation (dark rank 128). Hit = target continuation has lower mean NLL than the alternative.}
\label{tab:qwen-dark-hit-nd128}
\begin{tabular}{lccc}
\toprule
Dataset & no ablation & L16–19 & L20–23 \\
\midrule
NQ-Swap & 0.88 & 0.60 & 0.37 \\
ConflictQA & 0.98 & 0.96 & 0.85 \\
Cities & 1.00 & 1.00 & 1.00 \\
Sports & 0.95 & 0.92 & 0.94 \\
Celebrity & 0.96 & 0.94 & 0.95 \\
\bottomrule
\end{tabular}
\end{table}

\section{Datasets}
\label{sec:app-datasets}

\subsection{Contextual vs.\ parametric}
\label{sec:app-datasets-ctx-param}

Table~\ref{tab:nqswap-conflictqa-examples} illustrates the contextual conflict datasets used in Section~\ref{sec:empirical-study}: NQ-Swap and ConflictQA supply a passage whose injected answer contradicts world knowledge, so faithful answering requires following the context. We use 400 examples from each dataset.
For both datasets, we obtain each model's parametric response by querying it without the conflicting passage, then compare that continuation against the context-aligned answer under NLL.
Table~\ref{tab:factual-qa-examples} shows the complementary parametric knowledge datasets (Cities, Celebrities, Sports): closed-book yes/no questions with no passage, so answers must come from the model's stored knowledge. Cities contains 400 questions, while Celebrities and Sports contain 200 samples each.

\begin{table*}[t]
  \centering
  \small
  \setlength{\tabcolsep}{4pt}
  \begin{tabular}{@{}>{\raggedright\arraybackslash}p{0.16\textwidth}
                   >{\raggedright\arraybackslash}p{0.40\textwidth}
                   >{\raggedright\arraybackslash}p{0.40\textwidth}@{}}
  \toprule
   & \textbf{NQ-Swap}
   & \textbf{ConflictQA} \\
  \midrule
  Question
    & Who was the actor who played Crocodile Dundee?
    & Who is the author of \emph{How to Stop Worrying and Start Living}? \\
  \addlinespace
  Context
    \newline{\footnotesize(incorrect answer injected)}
    & \textbf{Gyanendra of Nepal}, AM (born 8 October 1939) is an Australian
      comedian, actor and television presenter. He was nominated for the
      Academy Award for Best Original Screenplay and won the Golden Globe
      Award for Best Actor\,---\,Motion Picture Musical or Comedy for his
      performance as outback adventurer Michael ``Crocodile'' Dundee in
      \emph{Crocodile Dundee} (1986), the first in the Dundee film franchise.
    & \textbf{Herbert Croly} is indeed the author of \emph{How to Stop Worrying
      and Start Living}. This can be confirmed by looking at the book's cover
      page, which clearly states ``By Herbert Croly'' as the author.
      Additionally, a quick search on reputable book-selling websites such as
      Amazon or Barnes\,\&\,Noble will also show that Herbert Croly is credited
      as the author of this book. Furthermore, Croly is a well-known author and
      political philosopher, having also written the influential book
      \emph{The Promise of American Life}. \\
  \addlinespace
  Injected incorrect answer
    & Gyanendra of Nepal
    & Herbert Croly \\
  \addlinespace
  Parametric answer
    & Paul Hogan
    & Dale Carnegie \\
  \bottomrule
  \end{tabular}
  \caption{Illustrative knowledge conflicts from NQ-Swap and ConflictQA.
  In both datasets the provided context has been altered so that it supports an
  \textbf{incorrect} answer (bolded in the passage).}
  \label{tab:nqswap-conflictqa-examples}
  \end{table*}

\begin{table*}[t]
  \centering
  \small
  \begin{tabular}{@{}lll@{}}
  \toprule
  \textbf{Dataset} & \textbf{Example question} & \textbf{Answer} \\
  \midrule
  Cities
    & Is the city of Tokyo in Japan?
    & Yes \\
    & Is the city of Tokyo in Pakistan?
    & No \\
  \addlinespace
  Celebrities
    & Is Leonardo DiCaprio an Academy Award-winning actor?
    & Yes \\
    & Is Rihanna from Brazil?
    & No \\
  \addlinespace
  Sports
    & Was Michael Jordan a famous basketball player?
    & Yes \\
    & Was Usain Bolt a famous swimmer?
    & No \\
  \bottomrule
  \end{tabular}
  \caption{Example yes/no questions from the Cities, Celebrities, and Sports
  factual probes. Each dataset mixes true and false statements; the model is
  asked to answer \emph{Yes} or \emph{No}.}
  \label{tab:factual-qa-examples}
  \end{table*}

\subsection{RAG-style benchmarks}
\label{sec:app-datasets-rag}

To evaluate the effectiveness of our approach across diverse RAG scenarios, we conduct experiments on three established English-language benchmarks spanning both question answering (QA) and summarization. These datasets present varying levels of difficulty for context grounding and hallucination detection.

\paragraph{MS MARCO \& CNN/DM.} We utilize the long-form QA (MS MARCO) and standard summarization (CNN/DM) tasks from the RAGTruth benchmark~\cite{niu2024ragtruth}. These subsets provide high-quality annotated data specifically designed for evaluating hallucinations in retrieval-augmented generation.

\paragraph{XSum.} We also include the Extreme Summarization dataset (XSum)~\cite{narayan2018don}. Because XSum requires highly abstractive, single-sentence summaries, models are notoriously prone to hallucination here. For XSum, both the model generations and hallucination annotations are adopted directly from \citet{kuleshova2026intrygue}. The XSum statistics are available in Table~\ref{tab:dataset_stats}.

\begin{table}[h!]
  \centering
  \caption{Hallucination statistics for XSum (annotations from~\citet{kuleshova2026intrygue}).}
  \begin{tabular}{l c c}
  \toprule
  Model & Total & Hal. Rate (\%) \\
  \midrule
  Mistral-7B & 749 & 39.4 \\
  Llama-2-7B & 746 & 34.9 \\
  Llama-2-13B & 730 & 32.3 \\
  Llama-3.1-8B & 650 & 28.9 \\
  Qwen3-8B & 750 & 24.3 \\
  gemma-3-4b & 650 & 31.3 \\
  \bottomrule
  \end{tabular}
  \label{tab:dataset_stats}
  \end{table}

\section{Main Results: implementation details}

In this section, we describe the key implementation choices:

\begin{itemize}

\item For \textsc{DarkLens}, hyperaparameters are selected on the validation set by maximizing AUROC:
we fix $\alpha=1$ and search $\mathrm{top}\text{-}k\in\{1,\ldots,L\}$, $\beta\in\{0.1,0.2,\ldots,2.0\}$, $\mathrm{null\text{-}dim}\in\{8,16,32,64,128,256\}$, and LM layer $\ell^*\in\{0,\ldots,L-1\}$, where $L$ is the number of model layers.
\item For ReDeEP, hyperaparameters are selected on the validation set by maximizing AUROC: we fix $\alpha=1$ fixed and search over $\mathrm{top}\text{-}k\in\{1,\ldots,L\}$, $\mathrm{top}\text{-}n\in\{1,\ldots,N\}$, and $\beta\in\{0.1,0.2,\ldots,2.0\}$, where $L$ is the number of layers and $N$ the number of copy heads.
\item For LUMINA we fix the cosine kernel and mixing weight $\lambda=0.5$ (following the original paper), with MMD computed over the top-$100$ tokens.
\item For EigenScore, we used the representation of the last token to embed sentences, as suggested by~\cite{chen2024inside}. We extracted outputs from the 16th layer, as middle layers have been shown to contain the most factual information~\cite{sky2024androids, azaria-mitchell-2023-internal}.
\item  For RAUQ, we performed the $\alpha$ search in $[0.0, 0.1, 0.2, \dots, 1.0]$ range.
\item All methods based on multiple generations employed $5$ additional model generations.

\end{itemize}

All experiments used 5-fold cross-validation. The train/val/test split fractions were set to 0.4/0.4/0.2, respectively. We used NVidia L40 and NVidia H100 GPUs; reproducing all reported results requires approximately 60-80 GPU hours.

\section{Use of scientific artifacts}

The RAGTruth and XSum benchmarks we use are publicly available under the MIT license, which permits their use for research purposes. The annotated XSum generations and labels were shared with us privately by the authors of~\cite{kuleshova2026intrygue}. Llama-2 models are released under the Llama 2 Community License Agreement. Mistral-7B and Qwen3-8B are available under the Apache 2.0 license. Gemma-3-4b-it is available under the Gemma Terms of Use. Llama-3.1-8B-Instruct is available under the Llama 3.1 Community License. Our experiments use the 3.7.3 version of the TransformerLens library~\cite{nanda2022transformerlens} (MIT license). We use all artifacts in this work in accordance with their respective licenses.

\section{Potential risks}

\textsc{DarkLens} evaluates faithfulness~---~how strictly an output is grounded in the provided context~---~rather than real-world factual correctness. If a model is prompted with inaccurate, biased, or adversarial retrieved documents, it may faithfully reproduce those errors. In such cases, \textsc{DarkLens} will appropriately assign high confidence, which users risk misinterpreting as a guarantee of objective truth.

\section{Use of AI} 
The text of this paper was proofread and refined for clarity using Gemini 3.8 Flash. Cursor models were used to assist the authors in writing the code for experiments and visualizing the results. The main idea and experimental design belong to the human authors of this work.

\end{document}